\documentclass[sigconf]{acmart}

\usepackage{booktabs}
\usepackage{array}
\usepackage{multirow}
\usepackage{listings}
\setcopyright{none}
\acmConference[AIAS+ 2026]{Chen Institute Symposium for AI Advancing Science and Society}
  {November 5--7, 2026}{San Francisco, CA, USA}
\acmYear{2026}
\acmDOI{}
\acmISBN{}

\begin{document}

\title{When Visual Critique Helps and Hurts: Structured Feedback for Scientific Plot Generation}

\author{Zhiyu Yang}
\affiliation{%
  \department{Erik Jonsson School of Engineering and Computer Science}
  \institution{The University of Texas at Dallas}
  \city{Richardson}
  \state{Texas}
  \country{USA}
}
\email{zhiyu.yang@utdallas.edu}

\begin{abstract}
Scientific visualization is a demanding setting for code-generating agents: a
plotting program may execute correctly while the rendered figure still violates
the user's intended chart type, visual encoding, layout, or annotations. We
reproduce a MatPlotAgent-style pipeline on MatPlotBench and study whether its
visual feedback stage can be made more explicit. Inspired by text-space visual
imagination, we replace free-form visual critique with a structured critic that
first describes the current figure and then converts the diagnosis into code
repair instructions. We evaluate direct generation, zero-shot chain-of-thought,
the reproduced agent pipeline, and the structured critic under both
reference-based and query-conditioned rubric evaluators. The structured critic
raises mean rubric scores across three backbones, although only the Gemini
3 Flash gain has a full-benchmark bootstrap interval excluding zero. Bucketed
and case-level analyses show that gains concentrate on
low-scoring or structurally flawed cases, while already strong figures are
vulnerable to over-editing. The evidence points to visual critique that is
explicit and calibrated.
\end{abstract}

\maketitle
\hypersetup{pdfauthor={Zhiyu Yang}}

\section{Introduction}

AI-assisted scientific work increasingly relies on agents that write and
execute code, inspect artifacts, and revise their outputs. Scientific
visualization exposes the limits of code-only feedback: a plotting script can
run successfully while the figure violates the requested data fields, chart
type, scales, encodings, labels, legends, layout, or annotations. Some errors
are visually subtle. A plot may look plausible but use the wrong hierarchy or
numeric values, so neither execution success nor visual plausibility alone
establishes that the request was satisfied. This makes plotting a useful
setting for examining how an agent interprets rendered output before editing
its code.

MatPlotAgent \cite{yang2024matplotagent} combines query expansion, code
generation, self-debugging, and visual feedback. Rather than stopping after
the code executes, it inspects the rendered figure and attempts to repair
visual mismatches. In our reproduction with a lightweight backbone, its
combined score is 71.56, versus 66.67 for direct generation and 67.65 for
zero-shot chain-of-thought (CoT) prompting.

Its free-form multimodal critic must perceive the image, compare it with the
query, diagnose discrepancies, and plan repairs in one response. It can
underdiagnose structural errors in difficult cases or propose unnecessary
edits to figures that already satisfy the request. Inspired by CapImagine
\cite{li2026capimagine}, we instead
require \textsc{VisualImagination}, describing the rendered figure and its
mismatches with the request, followed by \textsc{RevisionInstruction}, giving
code-level repairs. The first section acts as a textual record of what the
critic believes it sees; the second converts that diagnosis into changes for
the code agent. The interface makes it easier to inspect whether a bad repair
originated in perception or in the decision to intervene. For example, the
diagnosis can reveal that a chart has the requested form but maps the wrong
values, while the subsequent instruction reveals whether the critic asks for
a targeted data correction or a broader rewrite. Conversely, an accurate
description of an acceptable figure can still be followed by a harmful edit.

Across three generation backbones, structured critique improves full-benchmark
rubric means, substantially for Gemini 3 Flash but modestly for the other two.
Stratified and case analyses show gains on structurally flawed plots and losses
from over-editing strong ones. We additionally examine a prompt-development
held-out remainder, an independent-provider judge, and critic-call cost. The
central problem is therefore not only how to diagnose a plot, but when to
preserve it.

\begin{figure*}[t]
\centering
\includegraphics[width=\textwidth]{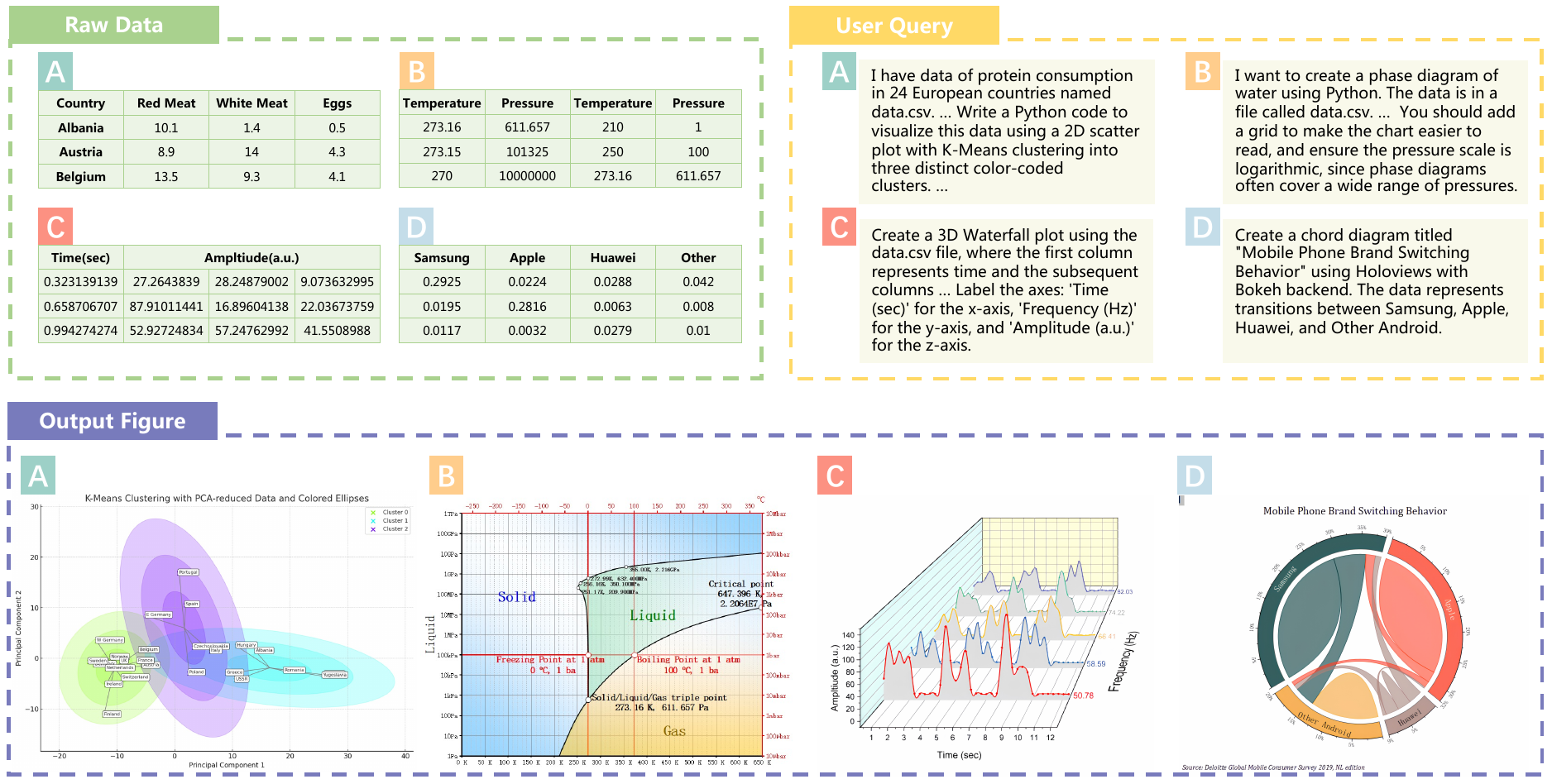}
\caption{Representative MatPlotBench tasks. Each pairs a visualization
request with raw data; the agent must generate a compliant figure.}
\Description{Examples of scientific plotting requests paired with their data and target plots.}
\label{fig:task_examples}
\end{figure*}

\section{Related Work}

\paragraph{Agent feedback.}
Tool-using agents \cite{nakano2021webgpt,yao2022webshop,wang2023voyager} and
textual self-revision methods \cite{madaan2023selfrefine,shinn2023reflexion}
show the value of feedback loops. They do not isolate how a multimodal critic
should translate visual evidence into safe code edits. Scientific plotting
makes that distinction observable because the code, rendered artifact,
feedback, and revised artifact can each be inspected.

\paragraph{Automatic visualization.}
Data2Vis \cite{dibia2018data2vis} maps input data to Vega-Lite visualization
specifications. LIDA \cite{dibia2023lida} proposes visualization goals from
a data summary, then generates and refines chart code for those goals.
MatPlotAgent \cite{yang2024matplotagent}
targets executable Python plots from raw data and detailed user queries,
including scales, projections, annotations, and layout. Unlike generic image
synthesis, the output must faithfully render data as well as satisfy visual
requirements.

\paragraph{Visual reasoning and explicit imagination.}
CapImagine \cite{li2026capimagine} uses explicit text-space imagination in
visual reasoning. We adapt this intermediate representation to plot repair,
where an apparently accurate description may still lead to an unnecessary or
overly broad code edit.

\begin{figure*}[t]
\centering
\includegraphics[width=0.88\textwidth]{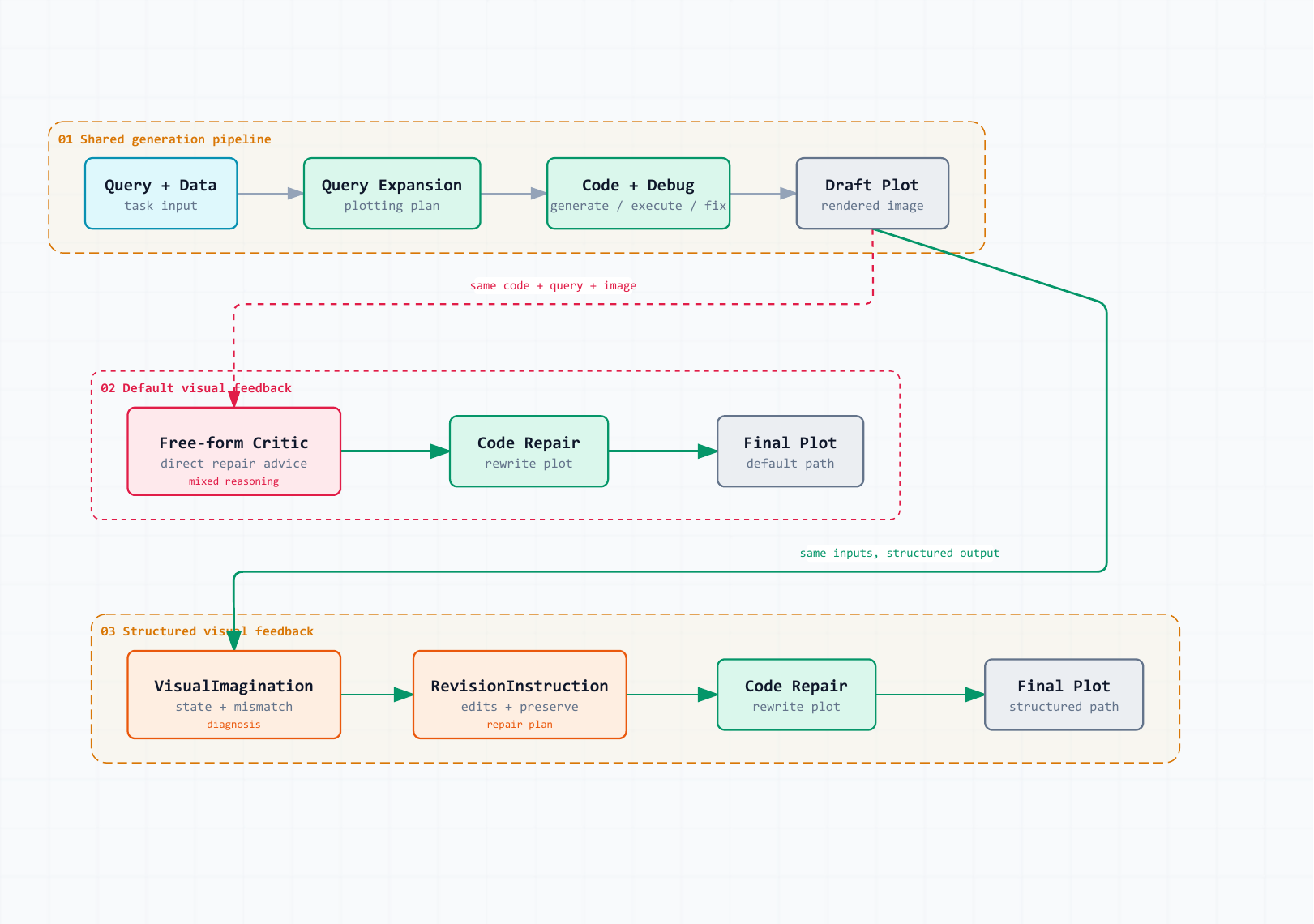}
\caption{Pipeline comparison. The MatPlotAgent visual critic directly
produces free-form repair feedback. Our structured critic adds an explicit
\textsc{VisualImagination} diagnosis before \textsc{RevisionInstruction},
creating an auditable text bottleneck before code repair.}
\Description{Both workflows expand the query, generate and debug code, and inspect the rendered plot. The baseline returns free-form feedback. The structured critic returns a visual diagnosis followed by code revision instructions.}
\label{fig:pipeline}
\end{figure*}

\section{Method}

\subsection{Task and Agent Framework}

Given a query and data file, the agent generates executable Python code whose
figure satisfies the requested encodings and layout. The reproduced pipeline
expands the query, generates and self-debugs code, then inspects the rendered
figure for repair. Within each setting, one backbone handles all stages;
self-debugging and visual refinement each allow at most three iterations.
Query expansion converts the original request into more explicit plotting
instructions. If generated code fails to execute, the code agent retries
using the execution error; after a draft image exists, the visual critic
supplies feedback for another code revision. We change the visual-feedback
interface without changing this stage sequence or its iteration limits.

\subsection{MatPlotAgent Visual Feedback}

The MatPlotAgent critic receives the current code, query, and rendered image,
then gives free-form, step-by-step repair advice. Its single response combines
perception, query comparison, discrepancy diagnosis, and repair planning. It
is instructed to compare plot type, data selection, labels, titles, colors,
and other visual elements with the request, and to improve clarity and
faithfulness. On a plausible but structurally wrong figure, this response
may be too vague to guide a repair; on an already adequate figure, it may
still encourage cosmetic changes. Because the response does not have to
separate observed properties from proposed edits, it can say a plot looks
mostly correct while leaving a wrong hierarchy or value mapping unaddressed.
The code agent then receives little indication of which part of the plot
must actually change.

\subsection{Structured Visual Feedback}

Our variant changes only that prompt, retaining the same inputs and one critic
call per refinement step. \textsc{VisualImagination} describes the figure,
separates satisfied from missing or incorrect requirements, and identifies
likely code causes; \textsc{RevisionInstruction} specifies repairs and what
should be preserved. The critic is asked to address data processing, plot
type, scales, labels, legends, layout, styling, annotations, and saving
behavior where relevant. The treatment pairs the two-part response format with an instruction to
preserve correct elements. The comparison estimates the effect of that full
prompt change. The structured prompt asks the critic to
identify what is already correct and to connect each proposed code change
to a visible or query-level discrepancy. A repair instruction that names the wrong data
mapping has a different effect from one that merely requests improved
readability. The complete prompts appear in the appendix.

\subsection{Evaluation}

\begin{figure*}[t]
\centering
\includegraphics[width=0.82\textwidth]{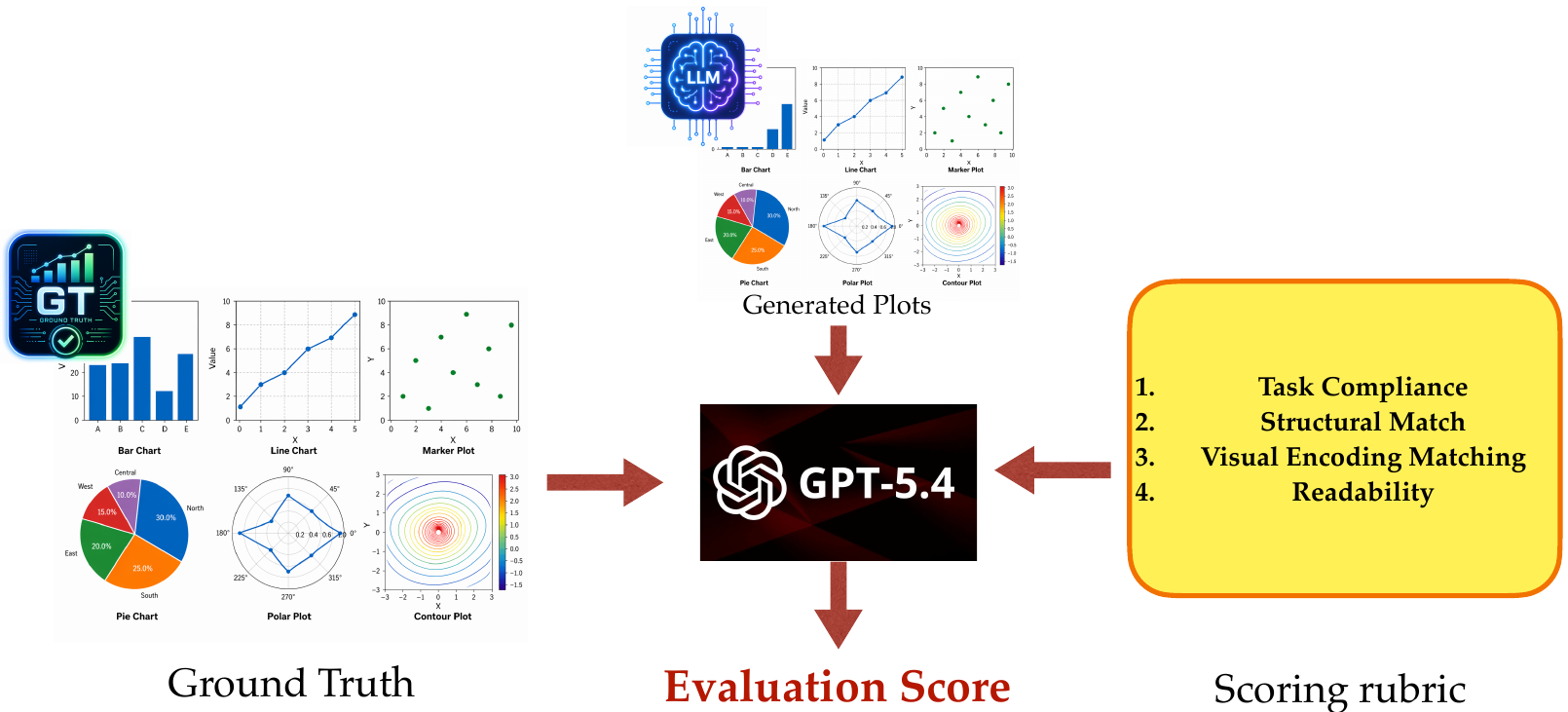}
\caption{Evaluation protocol. Beyond reference resemblance, the
query-conditioned rubric assesses task compliance, structure, visual encoding,
and readability.}
\Description{The reference-based judge compares the generated and reference figures. The rubric judge checks the rendered plot against the query using task compliance, structure, visual encoding, and readability criteria.}
\label{fig:evaluation_protocol}
\end{figure*}

The legacy evaluator scores resemblance to a reference figure. The
query-conditioned rubric judge scores task compliance, structure, encodings,
and readability (Fig.~\ref{fig:evaluation_protocol}). The
reproduction reports both and their average; the critic comparison uses rubric
score because it directly evaluates the request. A figure can resemble the
reference yet miss an explicit instruction, or look different while satisfying
the task; the two metrics therefore answer related but nonidentical questions.

\section{Experiments}

\subsection{Setup}

We evaluate on MatPlotBench \cite{yang2024matplotagent}, which contains 100
scientific visualization instances with a user query, raw data, and a
human-verified reference figure. We compare four settings: direct generation,
zero-shot chain-of-thought \cite{kojima2022zeroshot}, the reproduced
MatPlotAgent-style agent, and the CapImagine-inspired structured critic. We
evaluate three generation backbones: GPT-5.4-mini, Gemini 3 Flash, and Claude
Haiku 4.5. For automatic evaluation, we use GPT-5.4 and GPT-4o as multimodal
judges. All reported headline means are computed using the same evaluation
scorecard.

\subsection{Implementation Check}

Under the GPT-5.4 judge, the reproduced agent's combined score is 71.56,
versus 66.67 for direct generation and 67.65 for CoT
(Table~\ref{tab:reproduction}, appendix). Successful final images number
100, 95, and 93, respectively. The original evaluator can give a missing
output a small nonzero rubric score via an empty-image substitute; assigning
zero to every missing output yields combined means of 71.56, 66.23, and
66.69. Scores restricted to successfully scored outputs are 71.56, 69.71,
and 72.49 in the same order; one successful CoT image lacks a complete score
pair. Thus, the reproduced agent's advantage reflects both completion
reliability and figure quality. This check establishes a functioning
reproduction, not the critic effect, which uses paired completed examples.

\subsection{Main Results}

Structured critique improves full-benchmark rubric means for all three
backbones under both judges (Table~\ref{tab:main}). Gemini 3 Flash gains 8.41
points under GPT-5.4 and 7.70 under GPT-4o; the other gains are small. The
paired analyses below distinguish a consistent Gemini effect from those
uncertain changes rather than treating all positive means as equally strong
evidence.

\begin{table*}[t]
\caption{Rubric-score comparison between the MatPlotAgent (MPA) visual critic
and the structured critic. Gains are positive across all backbones and both
judges, but their magnitude varies substantially.}
\label{tab:main}
\centering
\footnotesize
\setlength{\tabcolsep}{2.8pt}
\begin{tabular}{lrrrrrr}
\toprule
 & \multicolumn{3}{c}{GPT-5.4 judge} & \multicolumn{3}{c}{GPT-4o judge}\\
\cmidrule(lr){2-4}\cmidrule(lr){5-7}
Backbone & MPA & Struct. & $\Delta$ & MPA & Struct. & $\Delta$\\
\midrule
GPT-5.4-mini     & 78.93 & 80.11 & +1.18 & 84.95 & 85.40 & +0.45\\
Gemini 3 Flash   & 70.02 & 78.43 & +8.41 & 74.93 & 82.63 & +7.70\\
Claude Haiku 4.5 & 76.87 & 77.49 & +0.62 & 84.84 & 85.40 & +0.56\\
\bottomrule
\end{tabular}
\end{table*}

\subsection{Leakage-Aware Held-Out Analysis}

MatPlotBench has no official development split. Prompt development used 30
examples selected by direct-generation rubric score (10 low, 10 middle, 10
high), not by structured-critic outcomes. The remaining 70 were not inspected
for prompt development, so we report them separately from the descriptive
full-100 result. On this remainder, Gemini gains
8.81 points under GPT-5.4 (95\% paired bootstrap CI [5.33, 12.38]) and 9.24
under GPT-4o ([4.13, 14.65]); the other changes are uncertain
(Table~\ref{tab:heldout70}). Some GPT-4o comparisons have fewer than 70
complete score pairs, as indicated in the table. This is a
baseline-stratified held-out remainder, not an independently supplied test
set.

\begin{table*}[t]
\caption{Rubric scores on the 70-example held-out remainder. Confidence
intervals use 50,000 paired, example-level bootstrap resamples. Some GPT-4o
rows contain fewer than 70 pairs because stored evaluator outputs are missing
for one GPT-5.4-mini example and three examples for each of the other two
backbones.}
\label{tab:heldout70}
\centering
\footnotesize
\setlength{\tabcolsep}{3.2pt}
\begin{tabular}{llrrrrl}
\toprule
Backbone & Judge & $n$ & MPA & Struct. & $\Delta$ & 95\% paired bootstrap CI\\
\midrule
GPT-5.4-mini & GPT-5.4 & 70 & 79.16 & 80.78 & +1.62 & [-1.65, 4.62]\\
GPT-5.4-mini & GPT-4o  & 69 & 84.47 & 84.58 & +0.11 & [-3.49, 3.64]\\
Gemini 3 Flash & GPT-5.4 & 70 & 69.53 & 78.34 & +8.81 & [5.33, 12.38]\\
Gemini 3 Flash & GPT-4o  & 67 & 73.11 & 82.35 & +9.24 & [4.13, 14.65]\\
Claude Haiku 4.5 & GPT-5.4 & 70 & 77.26 & 77.80 & +0.53 & [-1.69, 3.14]\\
Claude Haiku 4.5 & GPT-4o  & 67 & 84.01 & 84.61 & +0.60 & [-1.31, 2.63]\\
\bottomrule
\end{tabular}
\end{table*}

\subsection{Cross-Provider Judge Check}

We also froze a 30-case sample from the held-out remainder before new
evaluation: 10 per direct-score stratum, after excluding four cases without
complete stored artifacts. Selection used a fixed seed and no
structured-critic scores. Gemini 3.8 Flash judged the same stored figures
with the query-conditioned rubric. Its paired differences are $+8.42$ for
Gemini 3 Flash (95\% CI [4.64, 12.53]), $+2.99$ for GPT-5.4-mini, and
$-0.91$ for Claude (Table~\ref{tab:gemini38judge}). Thus, the large
Gemini gain also appears with a cross-provider judge, but gains are not
uniform.

\begin{table}[t]
\caption{Frozen-subset rubric evaluation by the cross-provider Gemini 3.8
Flash judge. Intervals use 50,000 paired bootstrap resamples. Win/tie/loss
counts are 15/4/11, 24/0/6, and 9/0/21, respectively.}
\label{tab:gemini38judge}
\centering
\footnotesize
\setlength{\tabcolsep}{3.5pt}
\begin{tabular}{lrrrrl}
\toprule
Backbone & $n$ & MPA & Struct. & $\Delta$ & 95\% paired bootstrap CI\\
\midrule
GPT-5.4-mini     & 30 & 68.03 & 71.03 & +2.99 & [0.68, 5.42]\\
Gemini 3 Flash   & 30 & 63.68 & 72.10 & +8.42 & [4.64, 12.53]\\
Claude Haiku 4.5 & 30 & 68.01 & 67.10 & -0.91 & [-3.61, 2.11]\\
\bottomrule
\end{tabular}
\end{table}

\subsection{Critic-Stage Computational Cost}

On those same 30 cases, both critics received identical stored code, query,
and draft image. The structured prompt used 59--90 more input tokens on
average but shorter responses, reducing mean total tokens, latency, and
estimated per-call cost for all three backbones
(Table~\ref{tab:critic_cost}, appendix). This measures one critic call, not
end-to-end workflow cost; the maximum call count is unchanged.

\section{Analysis}

Gemini's 0--20 and 20--40 legacy-score buckets gain 12.45 ($n=20$) and
16.65 ($n=26$) rubric points; GPT-5.4-mini gains 8.52 in its 20--40 bucket
($n=17$) but slightly regresses in higher buckets. Its 0--20 bucket has
only four cases, so that estimate is fragile
(Table~\ref{tab:buckets}, appendix). Gemini improves in every bucket,
whereas Claude, like GPT-5.4-mini, improves most on harder cases and is flat
or slightly worse on easier ones. Under GPT-5.4, paired rubric
win/tie/loss counts are 67/3/30 for Gemini, versus 49/5/46 and 49/4/47 for
GPT-5.4-mini and Claude. On all 100 paired differences, 20,000-resample
bootstrap 95\% intervals are [-1.57, 3.73] for GPT-5.4-mini, [5.28,
11.58] for Gemini, and [-1.24, 2.75] for Claude. Only Gemini's interval
excludes zero; the other mean gains are uncertain. We therefore interpret
the Gemini result as the primary quantitative finding, not as evidence that
the structured interface helps every backbone equally.

\subsection{Why Structured Critique Helps}

A separate qualitative review of 30 diagnostic cases found that successful
repairs often begin with plausible-looking figures whose values, hierarchy,
or geometry are wrong. For example, the structured critic made the
stacked-bar/table value mismatch in case 19 actionable
(Table~\ref{tab:case_studies}, appendix). Naming such errors gives the code
agent a specific repair target. In case 86, explicit diagnosis of radial
category ordering similarly improved the data-to-ring mapping. These are
cases where a generic assessment of visual plausibility could miss the
semantic mismatch even though the chart type looks appropriate. Other hard
cases have the right general appearance but the wrong geometry, such as a
requested 3D torus rendered as a vase-like surface. A critic that explicitly
names the missing structure can direct the code agent toward a geometric or
data-mapping repair instead of accepting the plot's surface resemblance.

The useful distinction is between recognizing the plot family and checking
the requested relationships inside it. A stacked bar chart can have the
right axes and colors but still encode values inconsistent with the supplied
table; a radial chart can contain rings without assigning the requested
categories to the correct rings. In these cases, advice to ``improve the
visualization'' is less actionable than an explicit account of what the
rendered figure shows and which query requirement it violates. The structured
critic makes that intermediate comparison visible to both the repair agent
and a human reader inspecting the failure.

\subsection{Why Structured Critique Hurts}

Detailed critique can also turn minor style concerns into unnecessary edits
to faithful plots. Case 14's already-close histograms degraded, while a local
sunburst styling issue became a brittle repair failure in case 88
(Table~\ref{tab:case_studies}). Better diagnosis alone does not ensure
better intervention. The repair model may change data distributions, view
angles, or label placement in response to low-impact observations. Such
changes can reduce fidelity or break a working implementation even when the
critic's description of the initial figure is reasonable. This is especially
visible in histograms and boxplots whose first draft already follows most
requirements: a detailed list of small concerns gives the repair model
permission to rewrite fragile data or layout code. In the sunburst example,
a local styling concern was escalated into a failed repair. The failure is
therefore often a mismatch between the importance of an observation and the
scope of the resulting edit, not simply a failure to see the image.

The contrast with the helpful cases is important. In a structurally wrong
plot, replacing a data mapping or geometry may be worth the risk. In an
already faithful plot, the same willingness to intervene becomes a liability:
a small concern about spacing or style can cause changes to code that also
controls data generation, axes, or view parameters. The revised plot may look
different without satisfying any additional requirement. Structured feedback
can make the observation sound more authoritative, but it does not by itself
tell the code agent how much of the existing implementation to preserve.

\subsection{Perception Error or Excessive Intervention?}

Of 17 degraded cases in that diagnostic set, manual classification attributes
13 mainly to intervention or repair, three to perception or diagnosis, and
one to an ambiguous requirement plus harmful repair. These selected cases
illustrate mechanisms, not population frequencies. All 10 easy-degraded
cases fall in the intervention/repair category, consistent with excessive
editing of figures that were already strong. The distinction matters: a
better visual description will not necessarily fix an over-eager editing
policy. In the three perception failures, the critic instead accepts a
plausible-looking but structurally incorrect chart, so the repair target is
wrong from the start. Separating these mechanisms helps explain why a more
explicit feedback format can improve hard examples and still hurt easy ones.
The two mechanisms also call for different remedies: better visual comparison
for the former, and a safer decision about whether and how broadly to edit
for the latter. The 13-to-3 imbalance in this selected set motivates examining
intervention calibration, but does not establish its prevalence on the full
benchmark.

\subsection{Diagnostic Test of a No-Revision Gate}

On the same 30 diagnostic cases with GPT-5.4-mini, a no-revision gate raised
the easy-degraded group's combined score from 76.34 to 79.08 but lowered the
hard-improved group's score from 74.80 to 63.38 and the overall mean from
63.72 to 62.19. An added edit-minimization instruction lowered the overall
mean further to 60.69. Among 10 hard-improved cases, the gate suppressed two
needed repairs and under-scoped three structural repairs. This is a diagnostic
negative result, not a new headline method: simple gating loses some of the
largest benefits. One additional failure came from repair execution and one
from the rendering environment. A useful gate would need to balance the risk
of modifying a good plot against the expected benefit of a substantial
repair, rather than make a binary decision from apparent visual quality
alone. The negative result is informative because the gate does solve part of
the problem it targets: it avoids some harmful edits on strong drafts. Yet
when the initial image has a serious but plausible-looking structural error,
accepting it or labeling the repair as minor can forfeit the improvement that
the structured critic otherwise achieved. Calibration must therefore account
for both the current plot's quality and the scale of the repair needed.

\section{Limitations}

Evaluation remains automatic: the cross-provider judge cannot replace human
assessment, and GPT-5.4 still evaluates a related generator. We also did not
test multiple generation seeds. These constraints limit the certainty of
small backbone-specific gains.

\section{Conclusion}

Structured visual critique makes the feedback stage of a scientific plotting
agent more interpretable and more useful on low-scoring or structurally flawed
cases. At the same time, detailed critique can over-edit already strong plots.
Reliable visualization agents need stronger visual reasoning and calibrated
intervention: knowing when to repair and when to preserve the current figure.

\bibliographystyle{ACM-Reference-Format}
\bibliography{aias2026_full_paper}

\clearpage
\appendix
\section{Additional Experimental Results}
\begin{table}[!htbp]
\caption{Bucket analysis under the GPT-5.4 judge. Buckets are defined by the
MatPlotAgent (MPA) workflow's legacy score, while reported cells are rubric
scores.}
\label{tab:buckets}
\centering
\footnotesize
\setlength{\tabcolsep}{2.8pt}
\begin{tabular}{llrrrr}
\toprule
Backbone & Bucket & $n$ & MPA & Struct. & $\Delta$\\
\midrule
\multirow{4}{*}{GPT-5.4-mini}
& 0--20  & 4  & 46.45 & 57.50 & +11.05\\
& 20--40 & 17 & 60.46 & 68.98 & +8.52\\
& 40--60 & 20 & 77.93 & 74.76 & -3.17\\
& 60+    & 59 & 86.79 & 86.67 & -0.13\\
\midrule
\multirow{4}{*}{Gemini 3 Flash}
& 0--20  & 20 & 42.53 & 54.97 & +12.45\\
& 20--40 & 26 & 59.88 & 76.53 & +16.65\\
& 40--60 & 13 & 80.05 & 83.95 & +3.90\\
& 60+    & 41 & 86.67 & 89.33 & +2.66\\
\midrule
\multirow{4}{*}{Claude Haiku 4.5}
& 0--20  & 5  & 32.96 & 41.84 & +8.88\\
& 20--40 & 24 & 67.10 & 70.23 & +3.12\\
& 40--60 & 16 & 73.70 & 72.47 & -1.23\\
& 60+    & 55 & 86.05 & 85.37 & -0.68\\
\bottomrule
\end{tabular}
\end{table}
\begin{table}[!htbp]
\caption{Implementation check on MatPlotBench with GPT-5.4-mini generation.
The reproduced codebase recovers the expected ordering: the MatPlotAgent-style agentic
pipeline outperforms direct generation and zero-shot CoT before we compare
visual-critic designs. Direct, CoT, and MatPlotAgent produce 95, 93, and 100
successful outputs. Their combined scores restricted to successfully scored
outputs are 69.71 ($n=95$), 72.49 ($n=92$; one successful output lacks a
complete score pair), and 71.56 ($n=100$), respectively. When every missing
output is instead assigned zero, the corresponding combined scores are 66.23,
66.69, and 71.56. The latter convention preserves the agent's advantage and
makes its completion-rate contribution explicit.}
\label{tab:reproduction}
\centering
\small
\setlength{\tabcolsep}{3.2pt}
\begin{tabular}{lrrrrrr}
\toprule
 & \multicolumn{3}{c}{GPT-5.4 judge} & \multicolumn{3}{c}{GPT-4o judge}\\
\cmidrule(lr){2-4}\cmidrule(lr){5-7}
Setting & Leg. & Rub. & Comb. & Leg. & Rub. & Comb.\\
\midrule
Direct  & 59.16 & 74.19 & 66.67 & 62.29 & 82.18 & 72.23\\
CoT     & 60.23 & 75.15 & 67.65 & 61.78 & 82.61 & 72.31\\
MatPlotAgent & 64.18 & 78.93 & 71.56 & 67.46 & 84.95 & 76.21\\
\bottomrule
\end{tabular}
\end{table}

\begin{table}[!htbp]
\caption{Cost of one visual-critic call on identical frozen inputs ($n=30$
per cell). Calls use the paper's models without an artificial output cap. Cost
is a standardized estimate using the provider-catalog token rates recorded on
September 17, 2026; actual billing can vary by endpoint or contract.}
\label{tab:critic_cost}
\centering
\scriptsize
\setlength{\tabcolsep}{2.4pt}
\begin{tabular}{lrrrrrr}
\toprule
& \multicolumn{2}{c}{Mean total tokens} & \multicolumn{2}{c}{Mean latency (s)} & \multicolumn{2}{c}{Est. cost/call (USD)}\\
\cmidrule(lr){2-3}\cmidrule(lr){4-5}\cmidrule(lr){6-7}
Backbone & MPA & Struct. & MPA & Struct. & MPA & Struct.\\
\midrule
GPT-5.4-mini     & 4776.9 & 4358.5 & 6.67 & 5.29 & 0.00850 & 0.00640\\
Gemini 3 Flash   & 3247.3 & 3153.3 & 6.67 & 5.84 & 0.00363 & 0.00316\\
Claude Haiku 4.5 & 4768.3 & 3734.9 & 12.97 & 8.82 & 0.01197 & 0.00644\\
\bottomrule
\end{tabular}
\end{table}

\begin{table}[!htbp]
\caption{Compact case studies from the GPT-5.4-mini qualitative read. The
$\Delta$ column reports differences in the combined score used for case
selection.}
\label{tab:case_studies}
\centering
\small
\setlength{\tabcolsep}{3.0pt}
\begin{tabular}{rllp{0.62\linewidth}}
\toprule
ID & Type & $\Delta$ & Main observation\\
\midrule
19 & helped & +39.35 & Correct form, wrong values; structured feedback made
the data/table mismatch actionable.\\
86 & helped & +48.60 & Wrong radial categories and ordering; explicit
diagnosis improved data-to-ring mapping.\\
14 & hurt & -26.60 & Already-close histograms; extra critique disturbed a
working implementation.\\
88 & hurt & -76.15 & Local sunburst styling issue became a brittle repair
failure.\\
\bottomrule
\end{tabular}
\end{table}
\section{Visual-Critic Prompts}

For reproducibility, we reproduce the two visual-critic prompts below. Runtime
placeholders are shown in brackets. Both conditions receive the same code,
query, and rendered image, and both use one visual-agent call per refinement
iteration.

\subsection{MatPlotAgent-Style Baseline}
\begin{lstlisting}[basicstyle=\ttfamily\scriptsize,breaklines=true]
SYSTEM_PROMPT = Given a piece of code, a user query, and an image of the current plot, please determine whether the plot has faithfully followed the user query. Your task is to provide instruction to make sure the plot has strictly completed the requirements of the query. Please output a detailed step by step instruction on how to use python code to enhance the plot.

USER_PROMPT = Here is the code: [Code]:
"""
{{code}}
"""

Here is the user query: [Query]:
"""
{{query}}
"""

Carefully read and analyze the user query to understand the specific requirements. Examine the provided Python code to understand how the current plot is generated. Check if the code aligns with the user query in terms of data selection, plot type, and any specific customization. Look at the provided image of the plot. Assess the plot type, the data it represents, labels, titles, colors, and any other visual elements. Compare these elements with the requirements specified in the user query. Note any differences between the user query requirements and the current plot. Based on the identified discrepancies, provide step-by-step instructions on how to modify the Python code to meet the user query requirements. Suggest improvements for better visualization practices, such as clarity, readability, and aesthetics, while ensuring the primary focus is on meeting the user's specified requirements.

Remember to save the plot to a png file. The file name should be """{{file_name}}"""
\end{lstlisting}

\subsection{Structured Critic}
\begin{lstlisting}[basicstyle=\ttfamily\scriptsize,breaklines=true]
SYSTEM_PROMPT = You are a visual reasoning assistant for matplotlib code repair. Given the current plotting code, the original user query, and the rendered plot image, perform explicit text-space imagination before proposing code changes.

First, reconstruct the important visual state of the chart in words: what is currently plotted, what appears missing, what seems incorrect, and which requirements from the query are or are not satisfied.

Then produce concrete code revision instructions that a code-generation model can follow to rewrite the plot.

Return your answer in exactly two sections with these headings:
1. VisualImagination
2. RevisionInstruction

The RevisionInstruction section should be specific, actionable, and focused on code changes needed to satisfy the user query.

USER_PROMPT = Here is the current plotting code: [Code]:
"""
{{code}}
"""

Here is the original user query: [Query]:
"""
{{query}}
"""

Analyze the query requirements, inspect the code, and examine the rendered plot image together.

Use text-space imagination to explicitly describe the current visual state and the intended target state before giving code edits.

In VisualImagination:
- Summarize what the figure currently shows.
- Identify mismatches between the query and the rendered chart.
- Mention likely root causes in the code when possible.

In RevisionInstruction:
- Give step-by-step instructions to modify the Python code.
- Preserve correct parts of the current code when possible.
- Be explicit about data processing, plot type, labels, legends, scales, layout, styling, annotations, and saving behavior when relevant.
- If the current chart already satisfies a requirement, say to keep that part unchanged.

Remember to save the plot to a png file. The file name should be """{{file_name}}"""
\end{lstlisting}

\end{document}